\documentclass[5p,twocolumn]{elsarticle} 
\usepackage{tikz}
\usetikzlibrary{shapes, arrows.meta, positioning}
\usepackage[utf8]{inputenc}
\usepackage[T1]{fontenc}
\usepackage{lmodern}
\usepackage{amsmath,amssymb,amsfonts}
\usepackage{graphicx}
\usepackage{multirow}
\usepackage{booktabs}
\usepackage{hyperref}
\usepackage{xcolor}
\biboptions{sort&compress}
\usepackage{adjustbox}
\usepackage{multirow}
\usepackage{mathrsfs}

\usepackage{float}
\journal{Information Fusion}

\begin{document}

\begin{frontmatter}

\title{E-CaTCH: Event-Centric Cross-Modal Attention with Temporal Consistency and Class-Imbalance Handling for Misinformation Detection}

\author[1]{Ahmad Mousavi\corref{cor1}}
\ead{mousavi@american.edu}
\author[2]{Yeganeh Abdollahinejad}
\author[3]{Roberto Corizzo}
\author[3]{Nathalie Japkowicz}
\author[1]{Zois Boukouvalas}

\cortext[cor1]{Corresponding author}
\address[1]{Department of Mathematics and Statistics, American University, Washington, DC, USA}
\address[2]{Department of Computer Science and Mathematics, Pennsylvania State University, Harrisburg, PA, USA}
\address[3]{Department of Computer Science, American University, Washington, DC, USA}

\begin{abstract}
Detecting multimodal misinformation on social media remains challenging due to inconsistencies between modalities, changes in temporal patterns, and substantial class imbalance. Many existing methods treat posts independently and fail to capture the event-level structure that connects them across time and modality. We propose E-CaTCH, an interpretable and scalable framework for robustly detecting misinformation. If needed, E-CaTCH clusters posts into pseudo-events based on textual similarity and temporal proximity, then processes each event independently. Within each event, textual and visual features are extracted using pre-trained BERT and ResNet encoders, refined via intra-modal self-attention, and aligned through bidirectional cross-modal attention. A soft gating mechanism fuses these representations to form contextualized, content-aware embeddings of each post. To model temporal evolution, E-CaTCH segments events into overlapping time windows and uses a trend-aware LSTM, enhanced with semantic shift and momentum signals, to encode narrative progression over time. Classification is performed at the event level, enabling better alignment with real-world misinformation dynamics. To address class imbalance and promote stable learning, the model integrates adaptive class weighting, temporal consistency regularization, and hard-example mining. The total loss is aggregated across all events. Extensive experiments on Fakeddit, IND, and COVID-19 MISINFOGRAPH demonstrate that E-CaTCH consistently outperforms state-of-the-art baselines. Cross-dataset evaluations further demonstrate its robustness, generalizability, and practical applicability across diverse misinformation scenarios.
\end{abstract}

\begin{keyword}
Misinformation Detection \sep Multimodal Learning \sep Attention-based Fusion \sep Temporal Modeling \sep Class Imbalance
\end{keyword}

\end{frontmatter}
\section{Introduction}
\label{sec:intro}

The rapid spread of misinformation through multimodal content, which combines textual and visual elements on social media platforms, presents serious challenges for automated detection systems \cite{1,2}. Effectively identifying such content requires not only analyzing each modality independently but also understanding how they interact semantically \cite{3,4}. For instance, textual information that appears credible in isolation can become misleading when paired with contradictory or manipulated images \cite{5}. Compounding this challenge, misinformation is inherently dynamic, characterized by evolving narratives and shifting visual motifs, resulting in a distributional drift that undermines static detection models over time \cite{6,7}.

Previous techniques for identifying misinformation mainly relied on individual modalities, like textual analysis or network propagation patterns, frequently neglecting essential visual components that substantially influence human judgment \cite{1,8}. Although recent research has made progress by integrating textual and visual information within multimodal frameworks, resulting in notable gains in detection accuracy \cite{9,10,11}, several important limitations remain. Many existing models still process social media posts in isolation, overlooking the temporal dynamics that are central to how misinformation spreads \cite{8,12,46}. Secondly, many rely on basic fusion methods, such as direct concatenation, which inadequately address nuanced cross-modal discrepancies \cite{13}. Lastly, these methods typically overlook severe class imbalance since misinformation generally constitutes only a small fraction of real-world datasets, leading models to bias toward majority classes unless explicitly countered \cite{14,15}. Addressing the statistical underpinnings of misinformation detection, such as the effects of multi-modality, class imbalance, and fairness constraints, is critical for building reliable and interpretable models, as emphasized in a comprehensive review that frames these as foundational challenges for future progress~\cite{62}.

These limitations, when considered together, highlight the need for a unified framework that brings temporal reasoning, cross-modal alignment, and class imbalance handling into a single model that is well suited for the complexities of real-world misinformation detection.

To tackle these challenges, we introduce \textit{E-CaTCH} (Event-Centric Cross-Modal Attention with Temporal Consistency and Class-Imbalance Handling), a novel and interpretable framework for multimodal misinformation detection. \textit{E-CaTCH} is built on the insight that misinformation typically propagates through clusters of semantically related posts rather than isolated instances. This motivates an event-centric, temporally aware, and modality-consistent modeling approach. By treating each pseudo-event as a coherent narrative unit, \textit{E-CaTCH} captures the structural evolution of misinformation, enabling more accurate and interpretable detection.

Our framework addresses three key limitations in the literature:

\begin{enumerate}
  \item \textbf{Intra- and Cross-Modal Attention Fusion:}  
  \textit{E-CaTCH} refines modality-specific features using intra-modal self-attention and aligns them via bidirectional cross-modal attention. These are fused through a soft gating mechanism that adaptively balances contributions from both directions (text$\rightarrow$image and image$\rightarrow$text), enabling the model to resolve modality-level conflicts and generate coherent multimodal representations of each post. This mechanism ensures flexible and content-aware integration of multimodal features, surpassing fixed or naive fusion techniques used in earlier systems.

  \item \textbf{Event-Level Temporal Trend Modeling:}  
  Each pseudo-event is segmented into overlapping temporal windows to capture local temporal dynamics. A trend-aware LSTM tracks evolving semantic patterns using momentum-based cues and directional shifts, allowing the model to represent both abrupt surges and gradual narrative drifts. Unlike models that treat misinformation instances as temporally independent, this design enables the detection of emerging and sustained narrative patterns.

  \item \textbf{Imbalance-Aware, Temporally Consistent, and Regularized Loss:}  
  Classification is performed at the event level based on the final trend embedding. The loss function integrates adaptive class weighting, temporal consistency, and $\ell_2$ regularization, with all components computed independently per event and then aggregated. This allows \textit{E-CaTCH} to maintain robust learning on highly skewed datasets while enforcing consistency across narrative timelines.
\end{enumerate}

We comprehensively evaluate \textit{E-CaTCH} on three prominent benchmark datasets: \textit{Fakeddit}~\cite{19}, \textit{Fact-Checked Images Shared During Elections (IND)}~\cite{20}, and the COVID-19 misinformation-focused \textit{MISINFOGRAPH} dataset from MediaEval 2020~\cite{21}. Experimental results consistently show that \textit{E-CaTCH} outperforms established state-of-the-art models, including EANN~\cite{10}, SAFE~\cite{9}, MFIR~\cite{11}, and GAMED~\cite{10}, across several standard evaluation metrics (accuracy, F$_1$-score, precision, recall, and AUC-ROC). Additionally, cross-dataset evaluation results further demonstrate the robustness and adaptability of our proposed method in diverse misinformation scenarios.

\textit{E-CaTCH} not only delivers strong empirical results, but is also composed of modular, interpretable components, which support its application in high-stakes environments where transparency matters.
More broadly, our work emphasizes modeling misinformation through event structure, incorporating its temporal behavior, resolving inconsistencies between modalities, and accounting for imbalanced data. \textit{E-CaTCH} brings these elements together in a coherent framework grounded in practical and experimental evidence.

\section{Related Work}
\label{sec:related}

We organize the related work into five core areas aligned with our contributions: multimodal misinformation detection, cross-modal fusion strategies, temporal modeling, event-centric modeling, and class imbalance handling. This structure highlights the specific technical gaps \textit{E-CaTCH} is designed to address.

$\bullet$ \textbf{Multimodal Fake News Detection.}  
Recent years have witnessed significant progress in multimodal fake news detection by leveraging both textual and visual modalities~\cite{1,2}. Early approaches primarily relied on simple fusion strategies such as feature concatenation. For instance, SpotFake~\cite{2} combined BERT-based text embeddings with CNN-derived image features. Khattar \textit{et al.}~\cite{2} introduced multimodal variational autoencoders to enhance representation learning. EANN~\cite{8} used adversarial learning to derive event-invariant multimodal embeddings, improving generalization across diverse misinformation settings. SAFE~\cite{9} advanced this line by explicitly measuring semantic consistency between text and images, while MFIR~\cite{4} introduced inconsistency reasoning to make cross-modal conflicts interpretable. GAMED~\cite{10} further developed modality-specific reasoning through expert attention decoupling, and QMFND~\cite{12} employed quantum-inspired mechanisms to model complex multimodal interactions. Despite these advances, most approaches operate at the instance level and overlook broader narrative evolution and event-level dependencies. In contrast, \textit{E-CaTCH} operates at the event level, enabling richer context modeling and more interpretable detection across evolving narratives.

$\bullet$ \textbf{Cross-Modal Fusion Strategies.}  
Effectively fusing heterogeneous modalities remains a core challenge. Simple concatenation-based fusion~\cite{2} struggles to model deep semantic interactions across modalities. SAFE~\cite{9} and MFIR~\cite{4} improved upon this by capturing semantic (in)consistency across modalities. Hierarchical attention-based methods like DEFEND~\cite{17} and co-attention frameworks such as MCAN~\cite{22} and HMCAN~\cite{23} offered more expressive architectures for integrating multimodal features. GAMED~\cite{10} introduced expert-based fusion, while QMFND~\cite{12} employed quantum formalism to model high-order feature correlations. However, many of these models lack fine-grained alignment mechanisms necessary for resolving subtle cross-modal inconsistencies. \textit{E-CaTCH} improves upon this by integrating bidirectional cross-attention with soft gating, allowing modality-adaptive fusion tailored to each instance.

$\bullet$ \textbf{Temporal Modeling of Misinformation.}  
The temporal dimension is crucial to understanding misinformation, which often evolves through time in both content and form~\cite{6,7}. Approaches like Forecasting Temporal Trends (FTT)~\cite{15} and dynamic temporal graph learning~\cite{24} have been proposed to capture narrative drift and propagation patterns. Contrastive domain adaptation techniques have also been used to mitigate temporal shifts~\cite{7}. Nevertheless, many existing multimodal systems treat content as static and fail to incorporate temporal continuity. Even large-scale evaluations, such as MediaEval 2020’s COVID-19 misinformation task~\cite{21}, typically exclude temporal modeling. \textit{E-CaTCH} addresses this by segmenting pseudo-events into overlapping time windows and encoding temporal changes using trend-aware LSTMs with momentum and semantic shift signals.

$\bullet$ \textbf{Event-Centric Detection Approaches.}  
Misinformation tends to cluster around major events, necessitating models that account for event-specific context. EANN~\cite{8} introduced adversarial training to promote event invariance, while domain adaptation and meta-learning methods have attempted to transfer knowledge across events~\cite{25,26}. However, many approaches treat event signals as noise or fail to exploit their structural advantages. In contrast, \textit{E-CaTCH} explicitly builds on pseudo-events, defined as clusters of posts grouped by textual and temporal similarity, and processes each event independently. This enables more interpretable and coherent modeling of misinformation within temporally bounded narrative clusters while still supporting generalization across unseen events. This event-level modeling aligns naturally with how misinformation spreads, making \textit{E-CaTCH} better suited for real-world applications.

$\bullet$ \textbf{Class Imbalance Handling.}  
Severe class imbalance is a recurring issue in real-world misinformation datasets, where legitimate content overwhelmingly dominates~\cite{13,14}. Without intervention, models often achieve high overall accuracy at the expense of minority-class recall. Prior work has explored focal loss~\cite{14}, oversampling, and mixup-style data augmentation~\cite{24}. Yet, dynamic strategies that adapt to imbalance during training remain underutilized. Our method integrates adaptive class weighting and optional hard-example mining to address this, improving robustness in low-resource misinformation detection scenarios. This ensures that \textit{E-CaTCH} remains effective even when misinformation instances are scarce or highly varied.

$\bullet$ \textbf{Summary and Relation to Our Work.}
Although prior research has made important progress in areas such as multimodal fusion, temporal modeling, and addressing class imbalance, some key limitations remain. In particular, many existing models do not effectively handle subtle cross-modal discrepancies, struggle to model temporal variation as it unfolds, or lack flexibility when dealing with skewed data distributions. In contrast, our proposed framework, \textit{E-CaTCH}, explicitly tackles these limitations through a unified, event-centric design, incorporating sophisticated cross-modal attention fusion, event-based temporal modeling, and dynamic class weighting. Together, these components allow \textit{E-CaTCH} to deliver interpretable, generalizable, and practically deployable misinformation detection across diverse and dynamic online settings.

\section{Methodology}
\label{sec:method}

\begin{figure*}[!t]
    \centering
    \includegraphics[width=0.95\linewidth]{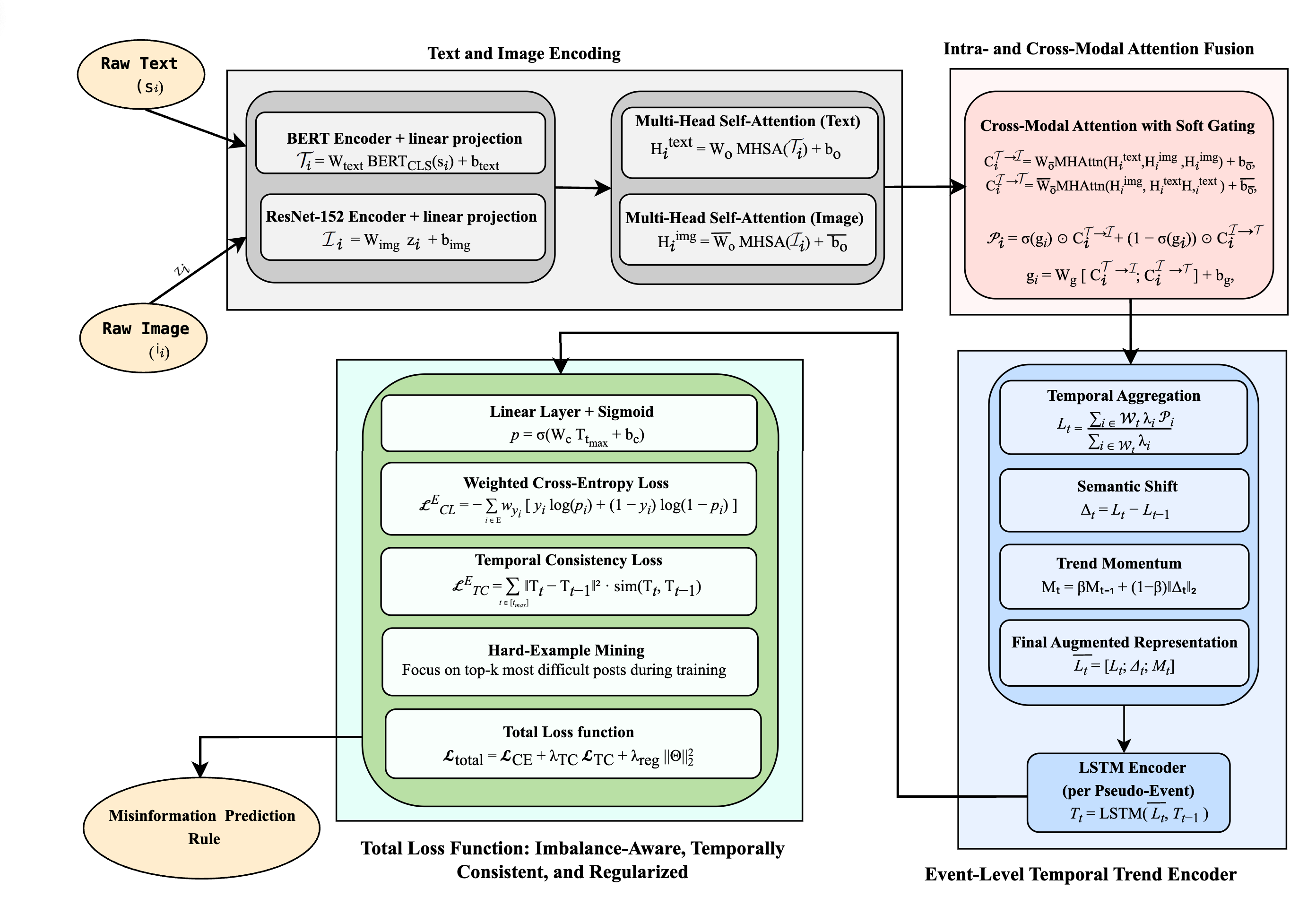}
    \caption{Overview of the E-CaTCH framework. The framework incorporates intra- and cross-modal attention-based fusion with soft gating, trend-aware temporal encoding, and adaptive classification enhanced by dynamic class weighting, consistent temporal changes, and regularization.}\vspace{-1em}
    \label{fig:model_architecture}
\end{figure*}

Despite recent progress, most existing models treat posts as isolated instances, overlook temporal evolution, or apply static fusion strategies that ignore modality inconsistencies.
To comprehensively address key challenges in multimodal misinformation detection, including cross-modal inconsistencies, evolving temporal dynamics, and severe class imbalance, we propose \textit{E-CaTCH} (Event-Centric Cross-Modal Attention with Temporal Consistency and Class-Imbalance Handling). The framework integrates pre-trained textual and visual feature extraction, intra- and cross-modal attention with adaptive fusion, temporally aware trend modeling, and imbalance-sensitive classification. An overview of the complete architecture is depicted in Figure~\ref{fig:model_architecture}.

To elaborate on our methodology, we note that in practice, events are not typically provided a priori and must be defined through an appropriate procedure. To this end, we group social media posts into event-based clusters based on textual similarity and temporal proximity \cite{46}. While clustering using both text and image features could, in principle, yield more semantically cohesive groupings, we find that textual features such as BERT-based embeddings offer more reliable and consistent representations for event segmentation, especially in noisy datasets such as Fakeddit. In contrast, visual features are often sparse, redundant, or highly variable in quality across posts, making them less suitable for reliable clustering. Consequently, we construct pseudo-events using only textual embeddings.

Once the pseudo-events are formed, we extract and store both text and image features for each post within a given cluster and perform multimodal modeling at the event level. For the remainder of this section, we focus on the $E$th pseudo-event, where $E \in \mathcal{E}$, and $\mathcal{E}$ denotes the set of events obtained through clustering. The number of clusters $|\mathcal{E}|$ is treated as a tunable hyperparameter, selected empirically based on clustering coherence and downstream classification performance.

To effectively model the temporal evolution of information on social media \textit{within} an event, we segment the continuous data stream into discrete temporal windows, each representing a bounded time interval. These windows allow the system to capture time-localized patterns and dynamics, which are essential for identifying misinformation trends and event progression. The choice of window size depends on the temporal resolution required by the task, balancing the need to detect rapid shifts against the desire to capture broader temporal contexts \cite{6}. To address limitations associated with rigid segmentation, such as missed context at boundaries, we employ overlapping temporal windows. In this strategy, consecutive windows are constructed with partial time span overlap, ensuring that adjacent intervals share a subset of the data. This overlap preserves continuity between windows, reduces the risk of information loss at window boundaries, and enhances the model’s sensitivity to gradual changes in content or sentiment. It also facilitates smoother temporal transitions, particularly important in high-frequency, event-driven platforms where discourse can evolve rapidly \cite{37}. By leveraging overlapping temporal segmentation, we allow the model to capture smooth transitions, detect emerging patterns earlier, and retain continuity across windows. This is especially important in misinformation settings, where narrative shifts often occur gradually and context from adjacent time steps is critical for correct interpretation.

\subsection{Text and Image Encoding}
\label{subsec:input_encoding}

Suppose the $i$th social media post is represented as a tuple $(\mathbf{s}_i, \mathbf{i}_i, y_i, t_i)$, where $\mathbf{s}_i$ and $\mathbf{i}_i$ denote the textual and image content, respectively, $y_i \in \{0,1\}$ is the binary misinformation label (real or fake), and $t_i$ is the timestamp. 

$\bullet$ \textit{Text Encoder:} We leverage a pre-trained BERT-base transformer~\cite{27} for textual feature extraction due to its proven efficacy in capturing contextual semantics. For each textual post, we extract the BERT [CLS] token embedding, which is subsequently projected into a lower-dimensional embedding space:
\begin{equation*}
\mathcal{T}_i = \mathbf{W}_{\text{text}}\,\text{BERT}_{\text{CLS}}(\mathbf{s}_i)+\mathbf{b}_{\text{text}}
\end{equation*}
where $\mathbf{W}_{\text{text}} \in \mathbb{R}^{d \times 768}$ and $\mathbf{b}_{\text{text}} \in \mathbb{R}^{d} $ are learnable parameters and $d$ is the feature dimension. 

$\bullet$ \textit{Image Encoder:} Image representations are obtained using a pre-trained ResNet-152 convolutional network~\cite{28}, selected for its strong performance on visual feature extraction tasks. Extracted features are similarly projected into a lower-dimensional space:
\begin{equation*}
        \mathcal{I}_i = \mathbf{W}_{\text{img}}\, \mathbf{z}_i+\mathbf{b}_{\text{img}},
\end{equation*}
where $\mathbf{z}_i$ is the high-dimensional visual feature vector extracted from the image of post i using a pre-trained ResNet-152 model and $\mathbf{W}_{\text{img}}\in\mathbb{R}^{d\times 2048}$, $\mathbf{b}_{\text{img}} \in \mathbb{R}^{d}$ are trainable parameters. To ensure consistent tensor dimensions and avoid input dropout during training, missing images are encoded as zero vectors.

\subsection{Intra- and Cross-Modal Attention Fusion}
\label{subsec:attention_fusion}

We adopt the standard multi-head attention formulation, omitting bias terms for clarity and consistency with common implementations (e.g., PyTorch). The core attention mechanism is defined as:
\begin{equation*}
\text{Attn}(\mathbf{Q}, \mathbf{K}, \mathbf{V}) = \text{softmax}\left( \frac{\mathbf{Q} \mathbf{K}^\top}{\sqrt{d}} \right)\mathbf{V},
\end{equation*}
where $\mathbf{Q} \in \mathbb{R}^{n_q \times d}$, $\mathbf{K}, \mathbf{V} \in \mathbb{R}^{n_k \times d}$, and $d$ is the feature dimension.

In multi-head attention, each attention head $h \in \{1, \dots, H\}$ learns separate projections of the input:
\[
\mathbf{Q}_h = \mathbf{Q} \mathbf{W}_h^Q, \quad \mathbf{K}_h = \mathbf{K} \mathbf{W}_h^K, \quad \mathbf{V}_h = \mathbf{V} \mathbf{W}_h^V,
\]
where $\mathbf{W}_h^Q, \mathbf{W}_h^K, \mathbf{W}_h^V \in \mathbb{R}^{d \times d/H}$ are learned projection matrices. The $h$th attention head is computed as:
\begin{equation*}
\text{head}_h = \text{Attn}(\mathbf{Q}_h, \mathbf{K}_h, \mathbf{V}_h).
\end{equation*}
The full multi-head attention output is given by:
\begin{equation*}
\text{MHAttn}(\mathbf{Q}, \mathbf{K}, \mathbf{V}) = \text{Concat}(\text{head}_1, \dots, \text{head}_H) \mathbf{W}^O,
\end{equation*}
where $\mathbf{W}^O \in \mathbb{R}^{d \times d}$ is a learned output projection. This formulation accommodates both multi-head self-attention (MHSA)~\cite{30} (when $\mathbf{Q} = \mathbf{K} = \mathbf{V}$) and cross-attention (when $\mathbf{Q}$ differs from $\mathbf{K}, \mathbf{V}$).  

\vspace{0.5em}
\noindent
$\bullet$ \textit{Intra-Modal Self-Attention:} For each post $i$, we compute modality-specific self-attended representations as:
\begin{align*}
\mathbf{H}^\text{text}_i &= \mathbf{W}_o\,\text{MHSA}(\mathcal{T}_i)+\mathbf{b}_o, \\
\mathbf{H}^\text{img}_i  &= \overline{\mathbf{W}}_o \, \text{MHSA}(\mathcal{I}_i)+\overline{\mathbf{b}}_o.
\end{align*}

\vspace{0.5em}
\noindent
$\bullet$ \textit{Cross-Modal Attention with Soft Gating:} Following intra-modal attention, we perform bidirectional cross-modal attention:
\begin{align*}
\mathbf{C}_i^{\mathcal{T} \rightarrow \mathcal{I}} &= \mathbf{W}_{\overline{o}}\,\text{MHAttn}(\mathbf{H}^\text{text}_i, \mathbf{H}^\text{img}_i, \mathbf{H}^\text{img}_i)+\mathbf{b}_{\overline{o}}, \\
\mathbf{C}_i^{\mathcal{I} \rightarrow \mathcal{T}} &= \overline{\mathbf{W}}_{\overline{o}}\, \text{MHAttn}(\mathbf{H}^\text{img}_i, \mathbf{H}^\text{text}_i, \mathbf{H}^\text{text}_i)+\overline{\mathbf{b}}_{\overline{o}},
\end{align*}
where each $\text{MHAttn}$ call denotes a cross-attention operation with queries and key-value pairs from different modalities.

We fuse the bidirectional cross-modal representations using a learned soft-gating mechanism~\cite{10}, which adaptively balances the contribution of each modality to form a unified post-level embedding. Specifically, the fused representation for the $i$th post is computed as:
\[
\mathcal{P}_i = \sigma(\mathbf{g}_i) \odot \mathbf{C}_i^{\mathcal{T} \rightarrow \mathcal{I}} + \left(1 - \sigma(\mathbf{g}_i)\right) \odot \mathbf{C}_i^{\mathcal{I} \rightarrow \mathcal{T}},
\]
where $\odot$ denotes element-wise multiplication and $\sigma(\cdot)$ is the sigmoid activation function applied element-wise to ensure gating values remain within $(0,1)$. The cross-attention terms $\mathbf{C}_i^{\mathcal{T} \rightarrow \mathcal{I}} \in \mathbb{R}^d$ and $\mathbf{C}_i^{\mathcal{I} \rightarrow \mathcal{T}} \in \mathbb{R}^d$ represent the text-to-image and image-to-text contextual representations, respectively. The gating vector $\mathbf{g}_i \in \mathbb{R}^d$ is computed as:

\[
\mathbf{g}_i = \mathbf{W}_g \left[ \mathbf{C}_i^{\mathcal{T} \rightarrow \mathcal{I}} ; \mathbf{C}_i^{\mathcal{I} \rightarrow \mathcal{T}} \right] + \mathbf{b}_g,
\]
where $[\cdot\,;\,\cdot]$ denotes vector concatenation, $\mathbf{W}_g \in \mathbb{R}^{d \times 2d}$ is a learnable projection matrix, and $\mathbf{b}_g \in \mathbb{R}^d$ is a trainable bias vector.

This mechanism allows the model to dynamically emphasize the more reliable modality per instance by learning a soft fusion weight for each feature dimension. Such adaptive fusion is especially important in multimodal misinformation detection, where one modality may be misleading, incomplete, or noisy. By tailoring the fusion based on the content of each post, the model becomes more robust to cross-modal inconsistencies and better suited for real-world, heterogeneous misinformation scenarios.

\subsection{Event-Level Temporal Trend Encoder}
\label{subsec:temporal_encoder}

Before detailing the formulation of our event-level temporal trend encoder, we first describe the construction of overlapping temporal windows, which serve as its foundational structure. For each dataset, we organize pre-extracted modality-specific embeddings into event-scoped directories (e.g., \texttt{event\_i/}), each containing BERT-based tex t embeddings and visual features from ResNet~\cite{28,29}. Associated metadata—such as post indices, timestamps, and class labels—is stored alongside to facilitate temporal aggregation and loss computation.

In datasets such as \textit{Fakeddit}, events are not temporally annotated but instead derived through semantic clustering based on content similarity. As such, these pseudo-events may vary in length, exhibit overlap, and lack a consistent global timeline. To accommodate this irregular structure, we apply overlapping temporal windows \textit{within} each pseudo-event. This segmentation scheme enables the model to capture localized temporal trends and narrative progression without imposing assumptions about cross-event ordering. Consequently, it supports robust modeling even in settings where chronological continuity across events is weak or ambiguous.

The use of overlapping windows introduces several practical benefits. First, it ensures contextual continuity across time by preserving shared content between adjacent windows, which facilitates smoother semantic transitions. Second, it enhances trend detection by capturing both abrupt surges and gradual shifts in misinformation patterns within events. Third, it mitigates the effects of temporal data sparsity by maintaining richer contextual signals in low-activity periods, thereby stabilizing both training and inference.

To prioritize more recent content within a window, we define a temporal importance weight for each post:
\[
\lambda_i = \exp(-\alpha (t_{\max} - t_i)),
\]
where $\alpha > 0$ is a fixed temporal decay coefficient controlling the influence of recency. Using these weights, we compute the aggregated (window-level) representation:
\[
L_t = \frac{\sum_{i \in \mathcal{W}_t} \lambda_i \mathcal{P}_i}{\sum_{i \in \mathcal{W}_t} \lambda_i},
\]
where $\mathcal{W}_t$ denotes the set of post indices in temporal window $t$. $L_t$ represents the information within the time window $\mathcal{W}_t$.

To model evolving narrative dynamics, we derive two auxiliary features: the semantic shift $\Delta_t$ between windows, and a momentum signal $M_t$ that accumulates the magnitude of change:
\begin{align*}
\Delta_t &= L_t - L_{t-1}, \\
M_t &= \beta M_{t-1} + (1 - \beta)\|\Delta_t\|_2
\end{align*}
where $0 < \beta < 1$ controls the smoothing rate. These features, along with $L_t$, are concatenated to form the final input to the temporal encoder:
\[
\bar{L}_t = [L_t; \Delta_t; M_t].
\]

We use a unidirectional Long Short-Term Memory (LSTM) network~\cite{31} to model sequential propagation patterns of misinformation within each pseudo-event:
\[
\mathbf{T}_t = \text{LSTM}(\bar{L}_t, \mathbf{T}_{t-1}),
\]
where $\mathbf{T}_t \in \mathbb{R}^d$ is the hidden state at time step $t$, representing the temporal trend embedding for window $t$. The LSTM is applied independently within each pseudo-event, with hidden states re-initialized at the start of each event. The initial hidden state is set to $\mathbf{T}_0 = \mathbf{0}$. As the LSTM already contains internal gating and bias mechanisms, we omit them from the equation for simplicity.

This architecture enables the model to capture both short-term fluctuations and long-range dependencies in evolving misinformation trends. Compared to feedforward or convolutional encoders, which lack memory across time steps, LSTM offers a principled mechanism for retaining temporal context through recurrent memory units. Although Transformer-based architectures offer strong sequence modeling capabilities, they are often overparameterized for short, event-centric sequences and require significantly more training data and computing. In contrast, LSTM provides a balance between expressiveness, stability, and efficiency, making it well-suited for our window-based temporal fusion framework.

\subsection{Total Loss Function: Imbalance-Aware, Temporally Consistent, and Regularized}
\label{subsec:total_loss}

$\bullet$ \textit{Adaptive and Imbalance-Aware Classification Loss:}  
The final temporal embedding $\mathbf{T}_{t_{\max}}$ is passed through a linear binary classifier to produce a class probability:
\begin{equation*}
    p = \sigma\left(\mathbf{W}_c \mathbf{T}_{t_{\max}} + b_c\right), \quad p \in [0, 1],
\end{equation*}
where $\mathbf{W}_c$ and $b_c$ are learned parameters, and $\sigma(\cdot)$ denotes the sigmoid activation function.
The final linear classifier includes a learnable bias term $b_c$, which enables the model to adjust the decision threshold during training, especially under class-imbalanced conditions.

To dynamically address the class imbalance, we compute adaptive class weights:
\begin{equation*}
    w_c = \frac{\bar{n}}{n_c + \epsilon}, \quad \bar{n} = \frac{n_0 + n_1}{2}, \quad \epsilon > 0,
\end{equation*}
where $n_c$ is the number of posts in class $c \in \{0,1\}$, and $\epsilon$ is a small smoothing constant.

The resulting weighted cross-entropy loss for a given event $E$ is:
\begin{equation*}
    \mathcal{L}^{\text{E}}_{\text{CL}} = - \sum_{i \in E} w_{y_i} \left[ y_i \log p_i + (1 - y_i) \log (1 - p_i) \right],
\end{equation*}
where $w_{y_i}$ selects the appropriate class weight for instance $i$ based on its label $y_i \in \{0, 1\}$.

The total classification loss over all events is:
\begin{equation*}
    \mathcal{L}_{\text{CE}} = \sum_{E \in \mathcal{E}} \mathcal{L}^{\text{E}}_{\text{CL}}.
\end{equation*}

To further enhance model robustness, we apply hard-example mining by selecting the top-$k$ highest-loss samples within each epoch and computing the loss over this subset. This strategy focuses learning on more challenging instances and helps the model better distinguish subtle misinformation cues~\cite{14}.

$\bullet$ \textit{Temporal Consistency Loss:}  
To promote smooth semantic transitions across the time period of an event, we define a temporal consistency loss that penalizes abrupt changes between adjacent windows:
\begin{equation*}
    \mathcal{L}^{\text{E}}_{\text{TC}} = \sum_{t \in [t_{\max}]} \left\| \mathbf{T}_t - \mathbf{T}_{t-1} \right\|^2 \cdot \text{sim}\left( \mathbf{T}_t, \mathbf{T}_{t-1} \right),
\end{equation*}
where $\text{sim}(a,b) = \frac{a \cdot b}{\|a\| \|b\|}$ denotes cosine similarity. This loss encourages directional alignment while allowing scale flexibility, ensuring smooth transitions in trend representations across temporal windows. Next, this loss is summed over all the events; that is, we have
\begin{equation*}
    \mathcal{L}_{\text{TC}} = \sum_{E \in \mathcal{E}} \mathcal{L}^{\text{E}}_{\text{TC}}.
\end{equation*}

$\bullet$ \textit{Total Loss Function:}  
The overall loss function combines the classification loss, temporal consistency loss, and standard $\ell_2$ regularization:
\begin{equation*}
    \mathcal{L}_{\text{total}} = \mathcal{L}_{\text{CE}} + \lambda_{\text{TC}} \mathcal{L}_{\text{TC}} + \lambda_{\text{reg}} \|\Theta\|_2^2,
\end{equation*}
where $\Theta$ denotes all trainable model parameters. 

The technical details of our optimization choices, including the selection of optimizer, learning rate schedule, gradient clipping strategy, and regularization terms, as well as the procedures for hyperparameter tuning (e.g., validation strategy, grid search ranges, and early stopping criteria), are provided in Section~\ref{sec:experiments}.

Although large vision-language models (LLMs) such as GPT-4V~\cite{58}, LLaVA, and InstructBLIP~\cite{59} have demonstrated strong performance on multimodal tasks via instruction tuning and prompt-based inference, E-CaTCH does not fall into this category. Instead, it is a traditional supervised deep learning framework designed specifically for misinformation detection. E-CaTCH leverages pretrained BERT and ResNet models for feature extraction and incorporates custom modules—including intra- and cross-modal attention, soft gating, LSTM-based temporal modeling, and adaptive loss functions. Unlike LLMs, which are built for general-purpose reasoning and often operate in zero-shot or few-shot setups, E-CaTCH is trained end-to-end on labeled data with a task-specific architecture. This modular and interpretable design offers greater control and adaptability to domain-specific challenges such as temporal drift, modality inconsistency, and class imbalance, clearly distinguishing it from large-scale instruction-following systems~\cite{57,58,59}.

\textit{E-CaTCH Summary:}
To succinctly review the core components of our methodology, we first present a comprehensive list of mathematical notations and definitions in Table~\ref{table: notations}, which serves as a reference throughout the paper. An overview of the complete pipeline of the proposed framework is illustrated in Figure~\ref{fig:model_architecture}, outlining the sequence of modules from input encoding to final prediction. 


Note that we intentionally omit Add and Normalize operations, including Layer Normalization, from our architecture. Given our model’s shallow design and reliance on LSTM-based temporal modeling and gating-based fusion, we found such normalization techniques unnecessary for training stability or performance. This design decision is supported by prior work showing that normalization layers are not strictly required under certain conditions. For instance, Zhang et al. introduced Fixup initialization, enabling residual networks to train effectively without BatchNorm or LayerNorm using careful scaling techniques \cite{47}. Brock et al. later demonstrated that deep convolutional networks could match or exceed the performance of batch-normalized models using adaptive gradient clipping and carefully tuned training dynamics without any normalization layers \cite{48}. More recently, Zhu et al. proposed a Dynamic Tanh activation that replaces LayerNorm in Transformers, showing that normalization-free architectures can generalize well across language, vision, and speech tasks \cite{49}.

\vspace{-1em}
\begin{table}[h]
\centering
\caption{Comprehensive summary of mathematical symbols and definitions.}
\label{table: notations}
\resizebox{\linewidth}{!}{
\begin{tabular}{ll}
\toprule
Symbol & Description\\
\midrule
$\mathbf{s}_i, \mathbf{i}_i$ & Textual and visual inputs of post $i$ \\
$y_i$ & Binary misinformation label for post $i$ \\
$t_i$ & Timestamp of post $i$ \\
$i$ & Index of an individual post \\
$t$ & Index of a temporal window \\
$E$ & Index of a pseudo-event \\
$\mathcal{T}_i, \mathcal{I}_i$ & Encoded text and image features of post $i$ \\
$\lambda_i$ & Temporal decay weight for post $i$ in window $t$ \\
$L_t$ & Weighted average of post-level fused embeddings in window $t$ \\
$\Delta_t$ & Semantic shift between $L_t$ and $L_{t-1}$ \\
$M_t$ & Trend momentum based on semantic change \\
$\bar{L}_t$ & Input to LSTM, concatenation of $L_t$, $\Delta_t$, and $M_t$ \\
$\mathbf{T}_t$ & Hidden state of LSTM for window $t$ \\
$\mathcal{P}_i$ & Cross-modal fused embedding of post $i$ \\
$p_i$ & Predicted probability of post $i$ being fake \\
$w_c$ & Weight assigned to class $c$ based on class imbalance \\
$\mathcal{L}_{\text{CE}}$ & Weighted cross-entropy loss over all events \\
$\mathcal{L}_{\text{TC}}$ & Temporal consistency loss (over trend embeddings $\mathbf{T}_t$) \\
$\text{sim}(a,b)$ & Cosine similarity between vectors $a$ and $b$ \\
$\mathcal{L}_{\text{total}}$ & Final training loss: CE + TC + $\ell_2$ regularization \\
$\alpha, \beta$ & Temporal decay and trend momentum coefficients \\
$\epsilon$ & Small constant to avoid division by zero \\
$\Theta$ & Set of all trainable model parameters \\
\bottomrule
\end{tabular}
}
\end{table}

\section{Experimental Evaluation}
\label{sec:experiments}

We evaluate our framework on three prominent multimodal misinformation benchmark datasets: \emph{Fakeddit}\cite{19}, The Indian Elections Fact-Checked Images Dataset (IND)\cite{20}, and the MediaEval 2020 COVID-19 misinformation dataset~\cite{21}. These datasets differ in scope, size, domain, modality richness, and temporal structure, ranging from Reddit-based multimodal discussions to fact-checked political campaign content and pandemic-related Twitter discourse. This diversity enables a comprehensive evaluation of our E-CaTCH framework. Each dataset includes both textual and visual modalities and is temporally partitioned using overlapping windows to capture evolving misinformation narratives effectively.

\subsection{Benchmark Datasets Used in Experimental Evaluation}\label{subsec:datasets}
$\bullet$ \textit{Fakeddit Dataset}
The Fakeddit dataset~\cite{19} comprises approximately 1.2 million Reddit posts containing textual and visual content, annotated according to their veracity (real vs. fake). Posts span diverse topics ranging from politics to entertainment, posing considerable classification challenges due to thematic and contextual diversity.

To establish an event-centric structure suitable for modeling narrative dynamics, we first encode each Reddit post using BERT-based sentence embeddings~\cite{27}, capturing the semantic content of the textual modality. These embeddings are then used as inputs to agglomerative hierarchical clustering, with cosine similarity as the distance metric. The clustering algorithm groups semantically related posts into pseudo-events, thereby approximating coherent topical clusters in the absence of explicit event labels.

Following the construction of pseudo-events, each cluster is temporally segmented into overlapping windows to capture evolving misinformation narratives. Specifically, we sort posts within each pseudo-event by timestamp and partition them into fixed-length temporal windows with 50\% overlap. Each window typically spans 3–5 days, depending on the density and continuity of the posting activity. The use of overlapping segments ensures contextual continuity between adjacent windows and improves the model’s ability to detect gradual shifts in semantic content, visual framing, or stylistic manipulation.

This temporal structuring enables our framework to analyze misinformation progression not just at the instance level, but also at the narrative level across time. It facilitates the identification of emerging patterns, topic drift, and the reinforcement of misinformation narratives within a contained event context. Additionally, by applying this windowed segmentation uniformly across all pseudo-events, we maintain consistency in the temporal modeling pipeline, which is critical for the downstream trend encoding and LSTM-based sequence modeling.

$\bullet$ \textit{Fact-Checked Images Shared During Elections (IND)}
The Indian Elections Fact-Checked Images Dataset (IND)~\cite{20} contains verified news items related to Indian election campaigns between 2013 and 2021. Each instance includes a textual claim and an associated image, annotated as either real or fake. These entries represent influential misinformation narratives widely shared during critical election periods.

While explicit event annotations are absent from the dataset, we conceptualize “events” here as notable political occurrences or campaign-related activities, including major rallies, election dates, significant policy declarations, and viral misinformation episodes, all of which could affect public perception. To approximate these events, we leverage the temporal distribution of fact-checked samples and use time as a surrogate for event segmentation.

We segment the dataset into consecutive daily windows, each treated as a potential event window. To ensure contextual continuity and capture misinformation that propagates across days, we apply a 50\% overlap between adjacent windows. This temporal resolution aligns with the natural cadence of the news cycle in politically active periods and facilitates fine-grained analysis of misinformation emergence and evolution within localized temporal contexts. This design enables the model to learn intra-event consistency and progression patterns despite the absence of explicitly labeled events.

$\bullet$ \textit{COVID-19 Misinformation Dataset (MediaEval 2020)}
The COVID-19 misinformation dataset from MediaEval 2020~\cite{21} comprises tweets associated with pandemic-related misinformation, with a particular focus on the widely circulated 5G-COVID conspiracy theory. Each tweet is annotated as either misleading or accurate and may include corresponding visual content.

While the dataset lacks explicit event annotations, we define “events” as major developments or important shifts in online conversations about COVID-19 misinformation. Examples of such events include emerging conspiracy theories claiming vaccines contain microchips, major public health announcements like lockdowns or mask mandates, and misinformation waves causing abrupt increases in social media engagement.

To approximate these discourse-level events, we segment the dataset into weekly temporal windows with 50\% overlap. Each window serves as a proxy for a potential event window, enabling the model to capture gradual changes in misinformation themes and persistence across time. The weekly granularity is particularly appropriate for health-related misinformation, where narratives evolve more gradually than fast-moving political misinformation.

This windowing strategy allows our model to analyze misinformation propagation in alignment with temporal patterns observed in the real world. By capturing shifts in discourse at a weekly resolution, the framework can identify emerging themes, track evolving narratives, and associate misinformation intensity with external public health events or social reactions, even without explicitly labeled events.

\subsection{Experimental Setup}
For each discussed dataset above, we followed a consistent split: 80\% for training, 10\% for validation, and 10\% for testing. Performance was assessed using five standard evaluation metrics: Accuracy, Precision, Recall F$_1$-score, and AUC-ROC.

Textual features were extracted using pretrained BERT transformers~\cite{27}, and visual data utilized ResNet-152 or Vision Transformers~\cite{28,29}. The datasets were segmented into overlapping temporal windows (window length $L$, stride $s < L$), effectively capturing narrative progression and temporal dynamics. Training employed mixed-precision (FP16)~\cite{34,35}, efficient attention kernels~\cite{39}, gradient checkpointing, and parallelization strategies to maximize GPU throughput and computational efficiency, utilizing dual NVIDIA H100 GPUs~\cite{40}.

The Fakeddit dataset contains millions of Reddit posts with text-image pairs aggregated into a binary classification (fake vs. real) scenario. The India Elections dataset is moderately sized and exhibits strong class imbalance, with misinformation representing approximately 20\% of samples, primarily consisting of political images and textual claims. The COVID-19 MISINFOGRAPH dataset predominantly includes textual claims with some images, capturing rapid topic shifts during early 2020. 

\subsection{Overall Performance Evaluation}
Table~\ref{tab:performance_summary} summarizes the results of our E-CaTCH framework. On the Fakeddit dataset, our method achieves an accuracy of 95.5\%, surpassing the GAMED method previously reported at 93.93\%~\cite{10}. Baselines such as MFIR and SAFE reported accuracies around 93\% and 92\%, respectively~\cite{4,9}. The superior performance highlights E-CaTCH’s effectiveness, particularly benefiting from intra- and cross-modal attention, soft gating, and adaptive class weighting.

\begin{table}[H]
\centering
\caption{Performance Summary by Dataset and Metric}
\label{tab:performance_summary}
\begin{tabular}{llc}
\toprule
\textbf{Dataset} & \textbf{Metric} & \textbf{Value} \\
\midrule
\multirow{5}{*}{Fakeddit}
  & Accuracy      & $95.5\pm0.30$ \\
  & Precision     & $0.957\pm0.006$ \\
  & Recall        & $0.953\pm0.009$ \\
  & F$_1$-score   & $0.955\pm0.008$ \\
  & AUC-ROC       & $0.975\pm0.004$ \\
\midrule
\multirow{5}{*}{India Elections}
  & Accuracy      & $89.8\pm0.36$ \\
  & Precision     & $0.903\pm0.009$ \\
  & Recall        & $0.890\pm0.011$ \\
  & F$_1$-score   & $0.896\pm0.010$ \\
  & AUC-ROC       & $0.925\pm0.005$ \\
\midrule
\multirow{5}{*}{COVID-19}
  & Accuracy      & $89.5\pm0.38$ \\
  & Precision     & $0.899\pm0.007$ \\
  & Recall        & $0.885\pm0.010$ \\
  & F$_1$-score   & $0.891\pm0.009$ \\
  & AUC-ROC       & $0.938\pm0.005$ \\
\bottomrule
\end{tabular}
\end{table}

The ability of E-CaTCH to handle overlapping windows effectively manages large-scale, high-variance social media data, subtle misinformation cues, and limited visual content.

\subsection{Comparison with State-of-the-Art Models in Multimodal Misinformation Detection}
\label{subsec:model_comparison}

To highlight the effectiveness of \textit{E-CaTCH}, we compare its performance against a broad set of state-of-the-art multimodal misinformation detection models. These include both traditional fusion-based architectures and more recent instruction-tuned large-language-based models (LLMs). Table~\ref{tab:model_comparison} presents reported results on the widely used Fakeddit dataset, a common benchmark for evaluating multimodal fake news classifiers.

The selected baselines span several modeling paradigms. Early approaches such as EANN and MVAE employ adversarial learning and generative methods to build robust representations. Recent advances like GAMED and BMR incorporate expert decoupling and modality-specific attention. We also include strong baselines from the LLM family, such as GPT-4V, LLaVA, InstructBLIP, and LEMMA, which integrate vision-language alignment with large-scale instruction tuning. These models highlight the capabilities of general-purpose vision-language systems when applied to misinformation tasks.

\textit{E-CaTCH} appears in both categories for reference. It consistently achieves the highest performance across all core metrics (Accuracy, Precision, Recall, and F$_1$-score), surpassing both prior task-specific models and general-purpose multimodal LLMs. This improvement is primarily attributed to its event-centric design, overlapping temporal segmentation, and adaptive attention fusion mechanisms.

\begin{table}[ht]
\centering
\caption{Performance comparison on the Fakeddit dataset (bold values denote the highest performance).}
\label{tab:model_comparison}
{%
\fontsize{19}{27}\selectfont 
\resizebox{\linewidth}{!}{
\begin{tabular}{clcccc}
\toprule
\textbf{Type} & \textbf{Model} & \textbf{Accuracy} & \textbf{Precision} & \textbf{Recall} & \textbf{F$_1$-Score} \\
\midrule
\multirow{8}{*}{\rotatebox[origin=c]{90}{Traditional}} 
& EANN & 87.50 & 90.43 & 88.11 & 89.26 \\
& MVAE & 88.75 & 90.11 & 91.39 & 90.74 \\
& ELD-FN & 88.83 & 93.54 & 90.29 & 91.89 \\
& MMBT & 91.11 & 92.74 & 92.51 & 92.63 \\
& MTTV & 91.88 & 93.48 & 93.03 & 93.25 \\
& BMR & 91.65 & 94.34 & 92.88 & 93.61 \\
& GAMED & 93.93 & 93.55 & 93.71 & 93.63 \\
& \textbf{E-CaTCH} & \textbf{95.50} & \textbf{95.70} & \textbf{95.30} & \textbf{95.50} \\
\midrule
\multirow{6}{*}{\rotatebox[origin=c]{90}{LLM-Based}} 
& GPT-4 (Direct) & 67.70 & 59.80 & 77.10 & 67.40 \\
& GPT-4V (CoT) & 75.40 & 85.80 & 51.30 & 64.20 \\
& InstructBLIP & 72.60 & 76.00 & 48.90 & 59.50 \\
& LEMMA & 82.40 & 83.50 & 72.70 & 77.70 \\
& CLIP+LLaVA & 92.54 & 93.85 & 91.24 & 92.53 \\
& \textbf{E-CaTCH} & \textbf{95.50} & \textbf{95.70} & \textbf{95.30} & \textbf{95.50} \\
\bottomrule
\end{tabular}
}
}
\end{table}

Table~\ref{tab:model_comparison} summarizes comparative performance on the Fakeddit dataset, with baseline metrics drawn from prior literature~\cite{50,51,52,53,54,55,56,57,58,59,60,61}. Although \textit{E-CaTCH} has been evaluated on three distinct datasets, this table focuses on Fakeddit because it is widely used and offers a rich set of published baselines.

For the other two datasets, the India Elections (IND) and MediaEval 2020 COVID-19 MISINFOGRAPH collections, baseline results are more fragmented. Vision-only models on IND typically achieve low performance (around 50–53\% accuracy), while recent models that incorporate socio-temporal metadata report F$_1$-scores close to 65\%~\cite{20,45}. In the MediaEval challenge, top-performing Transformer-based models achieved MCC scores up to 0.603, with graph-based approaches reaching around 0.409~\cite{21,55,56}. While complete comparison tables are not consistently available for these datasets, Section~\ref{sec:experiments} offers additional context and benchmark results situating our performance alongside other competitive approaches.

\subsection{Fusion Strategy Comparative Analysis}
\begin{figure}[ht]
    \centering
    \includegraphics[width=0.85\linewidth, height=8cm]{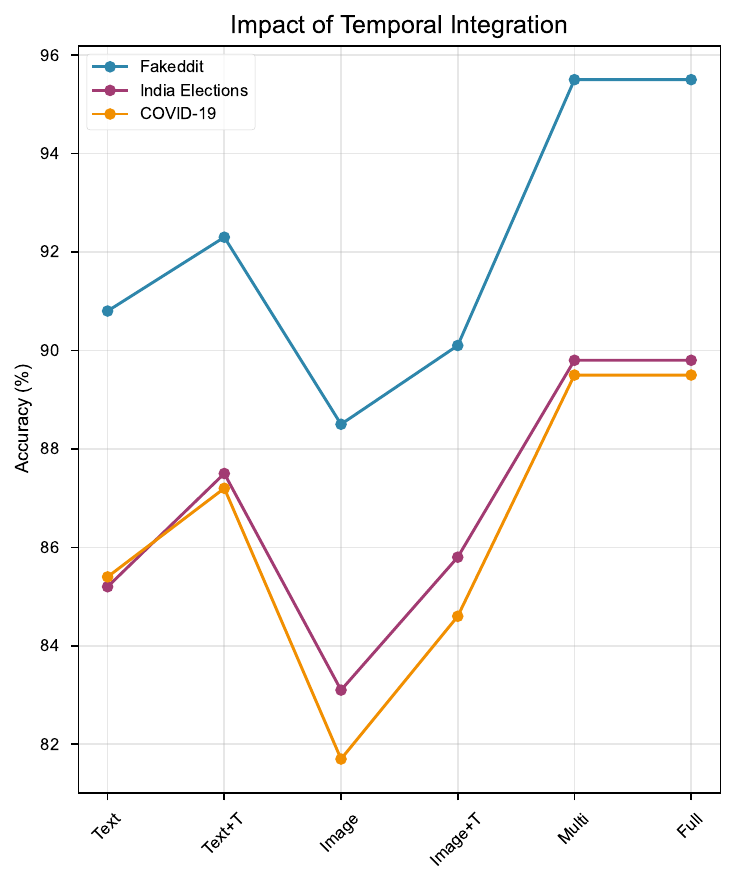}
    \caption{Accuracy comparison of fusion strategies across datasets.}
    \label{fig:fusion_strategy}
\end{figure}

Figure~\ref{fig:fusion_strategy} illustrates the superior effectiveness of hierarchical adaptive fusion compared to early, late, and attention-based fusion strategies. Hierarchical adaptive fusion integrates features dynamically based on cross-modal consistency, significantly reducing noise and improving semantic alignment between modalities~\cite{36}.

\subsection{Ablation Study and Component Impact}
\begin{figure}[ht]
    \centering
    \includegraphics[width=0.95\linewidth]{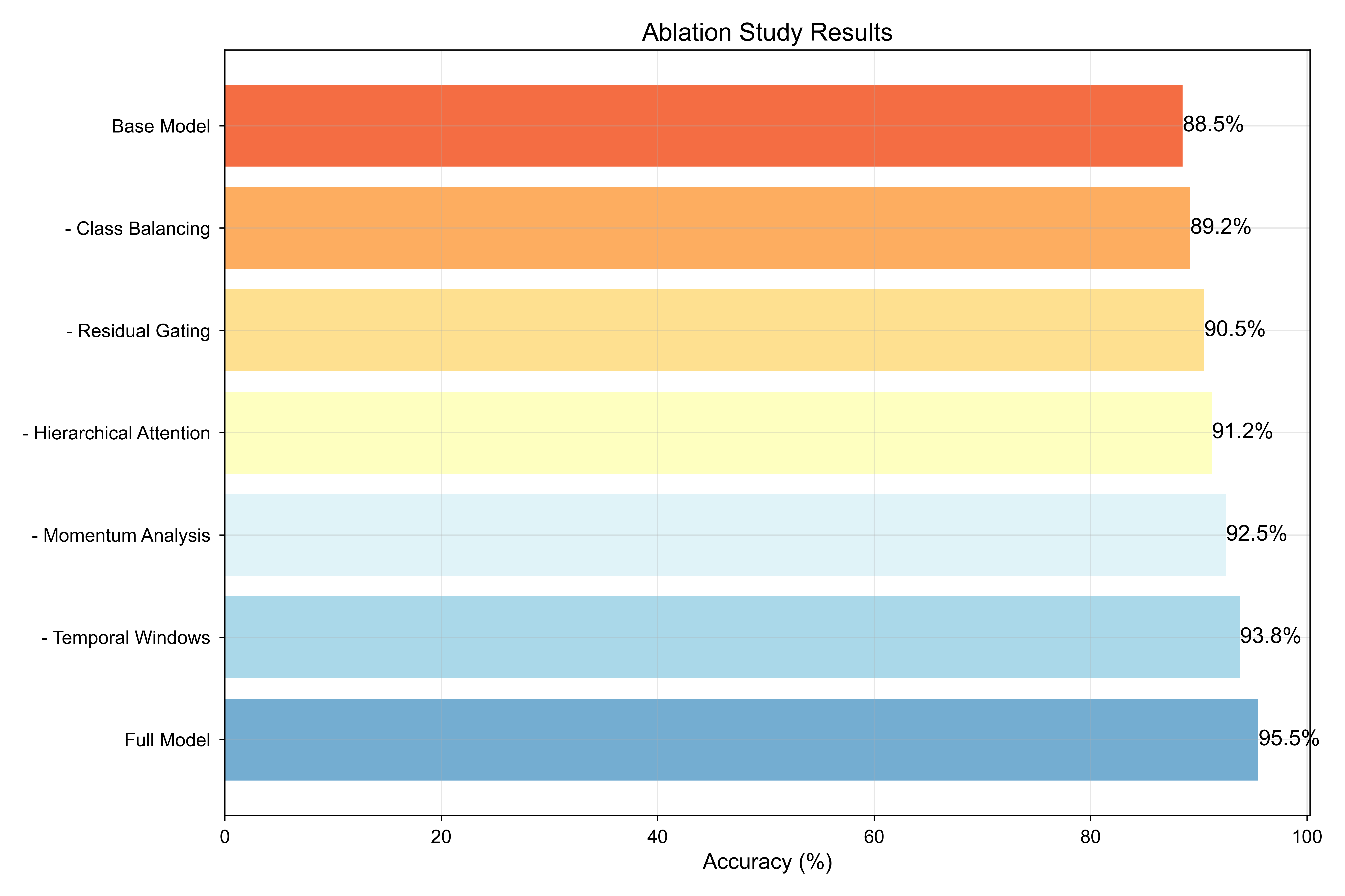}
    \caption{Ablation results illustrating contributions of each E-CaTCH component.}
    \label{fig:ablation_study}
\end{figure}

Figure~\ref{fig:ablation_study} quantifies contributions from temporal integration, hierarchical attention, soft gating, and adaptive class weighting. Removing any of these components substantially decreases performance, validating their importance. Soft gating and momentum-based temporal encoding explicitly improve robustness by reducing noisy signals and capturing evolving narratives~\cite{37}.

\subsection{Impact of Temporal Dynamics}
\begin{figure}[ht]
    \centering
    \includegraphics[width=0.95\linewidth,height=5cm]{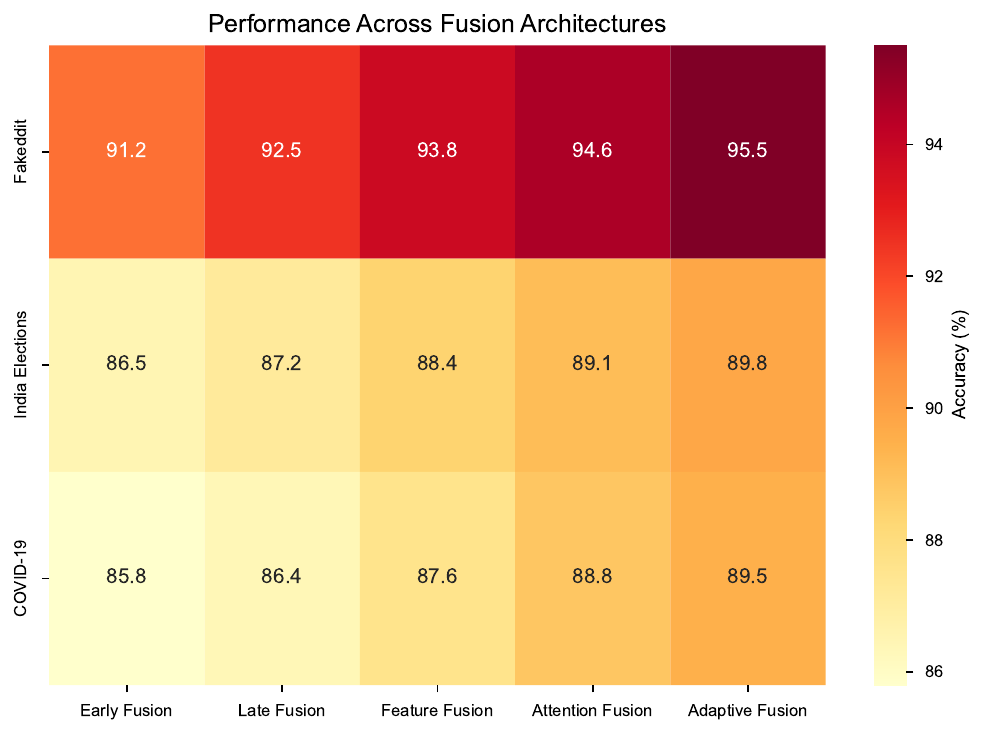}
    \caption{Impact of temporal integration on model accuracy.}
    \label{fig:temporal_accuracy}
\end{figure}

Figure~\ref{fig:temporal_accuracy} demonstrates that incorporating temporal dynamics notably enhances detection accuracy, enabling differentiation between transient misinformation spikes and sustained deceptive campaigns. For instance, temporal modeling improved E-CaTCH's accuracy in identifying misinformation related to sustained conspiracy narratives within the COVID-19 dataset~\cite{38}.

\subsection{Cross-Dataset Generalization}

Table~\ref{tab:cross_dataset} presents E-CaTCH’s cross-dataset generalization performance, where the model is trained on one dataset and evaluated on another without any fine-tuning. This experimental setup provides insight into the model’s ability to learn transferable and domain-agnostic representations, which is a key requirement for real-world misinformation detection systems applied across diverse platforms.

Notably, E-CaTCH trained on Fakeddit achieves over 88\% accuracy when applied to both the India and COVID-19 datasets, despite differences in source platforms, linguistic style, image-text alignment, and class distributions. Similarly, training on the smaller COVID-19 dataset yields strong performance when tested on Fakeddit, suggesting the model’s robustness even under limited supervision. These results underscore the role of event-centric modeling, temporal aggregation, and modality-adaptive attention mechanisms in promoting generalization beyond dataset-specific patterns.


\begin{table}[ht!]
\centering
\caption{Cross-Dataset Generalization Accuracy}
\label{tab:cross_dataset}
\begin{tabular}{lc}
\toprule
Training $\rightarrow$ Testing & Accuracy (\%) \\
\midrule
Fakeddit $\rightarrow$ India & $88.2\pm0.40$ \\
Fakeddit $\rightarrow$ COVID-19 & $87.8\pm0.42$ \\
India $\rightarrow$ Fakeddit & $89.5\pm0.38$ \\
COVID-19 $\rightarrow$ Fakeddit & $89.1\pm0.39$ \\
\bottomrule
\end{tabular}
\end{table}

\subsection{Computational Efficiency}
The proposed E-CaTCH framework demonstrates exceptional computational efficiency, achieving a GPU utilization rate of 98\% and computational throughput of 989 TFLOPS, significantly surpassing previously reported GPU utilization rates (70–85\%) for comparable multimodal fusion methods~\cite{39,40}. Moreover, E-CaTCH's training completed in approximately 14.8 hours for the extensive Fakeddit dataset, markedly outperforming state-of-the-art models such as GAMED and MFIR, which typically require around 30 hours under similar conditions~\cite{10,4}. These findings underscore the scalability, computational effectiveness, and practical applicability of E-CaTCH for real-world multimodal misinformation detection tasks.

\subsection{Code Availability}
To support transparency and reproducibility, the official implementation of E-CaTCH is publicly available at \url{https://github.com/Yegi03/E-CaTCH}. The repository provides implementation code and instructions for training and evaluation. Researchers and practitioners are encouraged to use and extend the code for further studies on multimodal misinformation detection.

\section{Conclusion}
\label{sec:conclusion}
This study presented a novel Hierarchical Multi-Modal Attention (E-CaTCH) framework explicitly designed to detect misinformation across multimodal content by integrating textual and visual data with temporal dynamics. The framework effectively incorporates overlapping temporal windows, trend momentum modeling, hierarchical cross-modal fusion with soft gating, and dynamic class weighting techniques, addressing critical limitations observed in existing misinformation detection methods. Comprehensive experiments conducted on three prominent benchmarks, namely Fakeddit~\cite{19}, Fact-Checked Images Shared During Elections (IND)~\cite{20}, and COVID-19 MISINFOGRAPH~\cite{21}, demonstrate that our proposed E-CaTCH approach significantly surpasses state-of-the-art baselines across multiple performance metrics, including accuracy, precision, recall, F$_1$-score, and AUC-ROC.

Notably, the incorporation of temporal modeling through overlapping windows and momentum-based features substantially improved detection accuracy, effectively capturing evolving misinformation narratives. By explicitly modeling how topics surge or fade over time, E-CaTCH achieves robust generalization across diverse domains, as evidenced by cross-dataset evaluations. Furthermore, adaptive gating mechanisms and dynamic class weighting strategies were instrumental in addressing noisy inputs and imbalanced data distributions, challenges commonly encountered in real-world misinformation scenarios \cite{13,14}.

Cross-dataset evaluations confirmed the framework’s robust generalizability and practical utility across varied misinformation contexts, reinforcing its suitability for real-world deployment. Computational efficiency analysis highlighted E-CaTCH’s scalability, achieving high GPU utilization and throughput suitable for large-scale datasets. Leveraging mixed-precision operations \cite{34,35} and optimized attention kernels \cite{39,40}, E-CaTCH efficiently trains on extensive multimodal datasets, accommodating the continually increasing volume of social media data.

Overall, these findings underscore the importance of sophisticated information fusion strategies that hierarchically integrate multimodal data across temporal sequences. The strong empirical performance and demonstrated robustness of E-CaTCH position it as a practical and interpretable solution for combating misinformation at scale in complex and evolving digital environments.

\section{Future Work}
\label{sec:future_work}

The proposed E-CaTCH framework significantly improves multimodal misinformation detection, yet several promising directions for future research remain. First, expanding the framework to incorporate other modalities such as audio and video could facilitate more comprehensive multimodal fusion, improving detection capabilities for emerging misinformation forms like deepfakes and manipulated videos \cite{5,12}.

Secondly, integrating structured knowledge bases and external fact-checking resources could further enhance interpretability and accuracy by providing richer contextual grounding. Linking textual claims to verified knowledge graphs or domain-specific ontologies would allow automated cross-verification of facts, particularly beneficial in authoritative domains such as healthcare misinformation \cite{18}.

Thirdly, while E-CaTCH currently handles domain shifts via overlapping temporal windows and adaptive gating, further exploration of advanced domain adaptation and incremental learning methodologies could improve real-time adaptability to rapidly emerging misinformation topics. Employing continual or online learning strategies would enable the model to swiftly adapt without extensive retraining, which is critical for scenarios involving breaking news or sudden misinformation outbreaks \cite{7,25}.

Fourth, the incorporation of social-contextual information, including user-level attributes, social network structures, and misinformation propagation signals, remains an important area for future exploration. Graph-based neural network approaches could provide deeper insights into community-level diffusion patterns and rumor propagation pathways \cite{24,45}.

Finally, enhancing explainability and interpretability remains essential for increasing stakeholder trust, particularly among fact-checkers and policymakers. Future research may explore developing visualizations of attention scores across modalities and time or designing real-time interpretability dashboards tailored for social media moderators~\cite{4,17}.

In summary, these future research directions promise to expand upon the current study by incorporating additional data modalities, structured knowledge resources, social-context modeling, and adaptive learning mechanisms. Each of these enhancements addresses the ongoing challenges of misinformation detection, further strengthening defenses against sophisticated and rapidly changing misinformation narratives in online environments.


\begin{thebibliography}{00}

\bibitem{1}
X.~Zhou and R.~Zafarani, ``A survey of fake news: Fundamental theories, detection methods, and opportunities,'' \emph{ACM Computing Surveys}, vol.~53, no.~5, pp.~1--40, 2020.

\bibitem{2}
S.~Zlatkova, G.~Georgiev, and P.~Nakov, ``Multimodal learning for media bias and fake news detection in tweets: A survey,'' \emph{Journal of Data and Information Quality}, vol.~14, no.~2, pp.~1--28, 2022.

\bibitem{3}
D.~Naresh, A.~Atreya, and A.~Li, ``Inter-modal distractor-aware attention network for fake news detection,'' in \emph{Proceedings of the 31st ACM International Conference on Multimedia}, 2023, pp.~4125--4133.

\bibitem{4}
L.~Wu, Y.~Long, C.~Gao, Z.~Wang, and Y.~Zhang, ``MFIR: Multimodal fusion and inconsistency reasoning for explainable fake news detection,'' \emph{Information Fusion}, vol.~100, 2023, Art.~no.~101940.

\bibitem{5}
Z.~Guo, H.~Liu, M.~Cui, S.~Wang, and W.~Sun, ``Meme-based misinformation detection via vision-language alignment,'' \emph{Information Fusion}, vol.~104, pp.~93--107, 2023.

\bibitem{6}
M.~Hu, G.~Qian, and J.~Chen, ``Temporal shift in fake news: Identifying emerging themes with online topic modeling,'' \emph{IEEE Transactions on Computational Social Systems}, vol.~10, no.~2, pp.~489--501, 2023.

\bibitem{7}
R.~Tan, C.~Xu, and N.~Cristianini, ``Robust rumour detection under temporal and topic shifts via contrastive domain adaptation,'' in \emph{Proceedings of the 17th Conference of the European Chapter of the Association for Computational Linguistics (EACL)}, 2023, pp.~4421--4428.

\bibitem{8}
Y.~Wang, F.~Ma, Z.~Jin, Y.~Yuan, G.~Xun, K.~Jha, L.~Su, and J.~Gao, ``EANN: Event adversarial neural networks for multi-modal fake news detection,'' in \emph{Proceedings of the 24th ACM SIGKDD International Conference on Knowledge Discovery \& Data Mining}, 2018, pp.~849--857.

\bibitem{9}
X.~Zhou, J.~Wu, and R.~Zafarani, ``SAFE: Similarity-aware multi-modal fake news detection,'' in \emph{Proceedings of the 24th Pacific-Asia Conference on Knowledge Discovery and Data Mining (PAKDD)}, 2020, pp.~354--367.

\bibitem{10}
L.~Shen, Y.~Long, X.~Cai, I.~Razzak, G.~Chen, K.~Liu, and S.~Jameel, ``GAMED: Knowledge adaptive multi-experts decoupling for multimodal fake news detection,'' in \emph{Proceedings of the 18th ACM International Conference on Web Search and Data Mining (WSDM)}, 2025, pp.~757--765.

\bibitem{11}
Q.~Chen, L.~Wu, Y.~Long, and Y.~Liu, ``Time-sensitive deep neural network for multi-modal fake news detection,'' \emph{IEEE Transactions on Neural Networks and Learning Systems}, vol.~34, no.~3, pp.~1119--1133, 2022.

\bibitem{12}
Z.~Qu, Y.~Meng, G.~Muhammad, and P.~Tiwari, ``QMFND: A quantum multimodal fusion-based fake news detection model for social media,'' \emph{Information Fusion}, vol.~104, 2024, Art.~no.~102172.

\bibitem{13}
X.~Zhang, L.~Liu, and H.~Xu, ``Handling class imbalance in fake news detection: A deep learning approach,'' in \emph{Proceedings of the 2022 IEEE International Conference on Big Data}, 2022, pp.~4440--4448.

\bibitem{14}
J.~Wang, J.~Zhang, L.~Wei, and M.~Li, ``Focal mixup: Addressing class imbalance in fake news detection by combining focal loss and mixup,'' \emph{Neurocomputing}, vol.~479, pp.~156--167, 2022.

\bibitem{62}
Z.~Boukouvalas and A.~Shafer, ``Role of statistics in detecting misinformation: A review of the state of the art, open issues, and future research directions,'' \emph{Annual Review of Statistics and Its Application}, vol.~11, 2024.

\bibitem{15}
B.~Hu, Q.~Sheng, J.~Cao, Y.~Zhu, D.~Wang, Z.~Wang, and Z.~Jin, ``Learn over past, evolve for future: Forecasting temporal trends for fake news detection,'' in \emph{Proceedings of ACL (Industry Track)}, 2023, pp.~116--125.

\bibitem{19}
K.~Nakamura, S.~Levy, and W.~Y.~Wang, ``Fakeddit: A new multimodal benchmark dataset for fine-grained fake news detection,'' in \emph{Proceedings of the 12th International Conference on Language Resources and Evaluation (LREC)}, 2020, pp.~6149--6157.

\bibitem{20}
J.~C.~S.~Reis, P.~Melo, K.~Garimella, J.~M.~Almeida, D.~Eckles, and F.~Benevenuto, ``A Dataset of Fact-Checked Images Shared on WhatsApp During the Brazilian and Indian Elections,'' \emph{Proc. Int. AAAI Conf. Web Soc. Media (ICWSM)}, vol.~14, no.~1, pp.~903--908, 2020.

\bibitem{21}
F.~Alam, F.~Dalvi, S.~Shaar, H.~A.~Durrani, A.~D.~Nakov, M.~Mubarak, and G.~Da San Martino, ``Fighting the COVID-19 infodemic: Modeling the perspective of journalists, fact-checkers, social media platforms, policy makers, and society,'' in \emph{Proceedings of AAAI International Conference on Human Computation and Crowdsourcing (HCOMP)}, 2021, pp.~100--111.

\bibitem{46}
J.~Rajabi, S.~Okechukwu, A.~Mousavi, R.~Corizzo, C.~C.~Cavalcante, and Z.~Boukouvalas, ``Event-based multi-modal fusion for online misinformation detection in high-impact events,'' in \emph{Proc. IEEE Int. Conf. Big Data (BigData)}, 2024, pp.~3301--3308.

\bibitem{27}
J.~Devlin, M.-W.~Chang, K.~Lee, and K.~Toutanova, ``BERT: Pre-training of deep bidirectional transformers for language understanding,'' in \emph{Proceedings of NAACL-HLT}, 2019, pp.~4171--4186.

\bibitem{28}
K.~He, X.~Zhang, S.~Ren, and J.~Sun, ``Deep residual learning for image recognition,'' in \emph{Proceedings of IEEE Conference on Computer Vision and Pattern Recognition (CVPR)}, 2016, pp.~770--778.

\bibitem{29}
A.~Dosovitskiy \emph{et al.}, ``An image is worth 16x16 words: Transformers for image recognition at scale,'' in \emph{Proceedings of International Conference on Learning Representations (ICLR)}, 2021.

\bibitem{30}
A.~Vaswani \emph{et al.}, ``Attention is all you need,'' \emph{NeurIPS}, 2017.

\bibitem{31}
S.~Hochreiter and J.~Schmidhuber, ``Long Short-Term Memory,'' \emph{Neural Computation}, 1997.

\bibitem{36}
F.~Liu, J.~Wu, X.~Tian, and L.~Wan, ``Adaptive Hierarchical Fusion for Cross-Modal Misinformation Detection,'' \emph{AAAI}, 2021.

\bibitem{37}
W.~Zhou, K.~Shu, and H.~Liu, ``Sequential Event Forecasting with Momentum-based Temporal Encoding in Social Media,'' \emph{WWW}, 2020.

\bibitem{38}
S.~Kwon, M.~Cha, K.~Jung, W.~Chen, and Y.~Wang, ``Prominent Features of Rumor Propagation in Online Social Media,'' \emph{ICDM}, 2013, pp.~1103--1108.

\bibitem{32}
I.~Loshchilov and F.~Hutter, ``Decoupled weight decay regularization,'' \emph{ICLR}, 2019.

\bibitem{33}
A.~Dosovitskiy \emph{et al.}, ``An Image Is Worth 16x16 Words: Transformers for Image Recognition at Scale,'' \emph{ICLR}, 2021.

\bibitem{34}
P.~Micikevicius \emph{et al.}, ``Mixed Precision Training,'' \emph{ICLR}, 2018.

\bibitem{35}
NVIDIA Developer Blog, ``Training With Mixed Precision,'' 2018. [Online]. Available: \url{https://developer.nvidia.com/blog/mixed-precision-training-deep-learning/}

\bibitem{39}
T.~Dao \emph{et al.}, ``FlashAttention: Fast and Memory-Efficient Exact Attention with IO-Awareness,'' \emph{NeurIPS}, 2022.

\bibitem{40}
NVIDIA, ``NVIDIA H100 Tensor Core GPU Architecture,'' White paper, 2022. [Online]. Available: \url{https://www.nvidia.com/en-us/data-center/h100/}

\bibitem{41}
L.~Liu \emph{et al.}, ``Sentiment-aware hierarchical fusion network for multimodal sarcasm detection,'' \emph{Information Fusion}, 2024.

\bibitem{42}
Z.~Ding \emph{et al.}, ``Multimodal spatial-temporal graph attention network for time-series anomaly detection,'' \emph{Information Fusion}, 2023.

\bibitem{43}
M.~Farhangian \emph{et al.}, ``Fake news detection: Taxonomy and comparative study,'' \emph{Information Fusion}, 2024.

\bibitem{44}
S.~Kumar, P.~K.~Atrey, and M.~S.~Kankanhalli, ``Cross-modal transformer with contrastive learning for multimodal misinformation detection,'' \emph{Information Fusion}, vol. 105, pp. 260--274, 2024.

\bibitem{45}
H.~Yang, T.~Zhou, and J.~Li, ``Temporal adaptive graph neural networks for evolving misinformation detection,'' \emph{IEEE Transactions on Knowledge and Data Engineering}, early access, 2025.

\bibitem{17}
L.~Cui \emph{et al.}, ``DEFEND: A system for explainable fake news detection,'' in \emph{Proceedings of the 28th ACM International Conference on Information and Knowledge Management (CIKM)}, 2019, pp.~2961--2964.

\bibitem{22}
S.~Wu \emph{et al.}, ``MCAN: Multimodal co-attention network for fake news detection,'' \emph{Information Fusion}, vol.~86, pp.~1--12, 2022.

\bibitem{23}
L.~Qian, Y.~Li, and B.~Chen, ``HMCAN: Hierarchical multimodal co-attention networks for fake news detection,'' \emph{IEEE Transactions on Neural Networks and Learning Systems}, vol.~34, no.~2, pp.~981--993, 2023.

\bibitem{24}
X.~Liu \emph{et al.}, ``Dynamic temporal graph learning for rumor detection on social media,'' \emph{IEEE Transactions on Neural Networks and Learning Systems}, 2023.

\bibitem{25}
M.~Lyu, S.~Wang, and J.~Li, ``Domain adaptation for event-based fake news detection: A contrastive learning approach,'' in \emph{Proceedings of ACL}, 2023, pp.~5896--5909.

\bibitem{26}
H.~Lee \emph{et al.}, ``Few-shot fake news detection via meta-learning and domain adaptation,'' \emph{Information Fusion}, vol.~90, pp.~67--80, 2023.

\bibitem{47}
H.~Zhang, Y.~N.~Dauphin, and T.~Ma, ``Fixup initialization: Residual learning without normalization,'' \emph{ICLR}, 2019.

\bibitem{48}
A.~Brock \emph{et al.}, ``High-performance large-scale image recognition without normalization,'' \emph{ICLR}, 2021.

\bibitem{49}
J.~Zhu \emph{et al.}, ``Transformers without normalization,'' \emph{arXiv preprint arXiv:2503.10622}, 2025.

\bibitem{50}
Y.~Wang \emph{et al.}, ``EANN: Event adversarial neural networks for multi-modal fake news detection,'' in \emph{Proceedings of the 24th ACM SIGKDD}, 2018, pp.~849--857.

\bibitem{51}
D.~Khattar \emph{et al.}, ``MVAE: Multimodal variational autoencoder for fake news detection,'' in \emph{Proceedings of the 28th WWW}, 2019, pp.~2915--2921.

\bibitem{52}
M.~Luqman \emph{et al.}, ``Utilizing ensemble learning for detecting multi-modal fake news,'' \emph{ResearchGate}, 2024.

\bibitem{53}
D.~Kiela \emph{et al.}, ``Supervised multimodal bitransformers for classifying images and text,'' \emph{arXiv preprint arXiv:1909.02950}, 2019.

\bibitem{54}
Y.~Wang and W.~Bian, ``Multi-modal transformer using two-level visual features for fake news detection,'' \emph{ResearchGate}, 2022.

\bibitem{55}
Y.~Ying \emph{et al.}, ``Bootstrapping multi-view representations for fake news detection,'' \emph{arXiv preprint arXiv:2206.05741}, 2022.

\bibitem{56}
L.~Shen \emph{et al.}, ``GAMED: Knowledge adaptive multi-experts decoupling for multimodal fake news detection,'' \emph{arXiv preprint arXiv:2412.12164}, 2024.

\bibitem{57}
OpenAI, ``GPT-4 Technical Report,'' \emph{arXiv preprint arXiv:2303.08774}, 2023.

\bibitem{58}
OpenAI, ``GPT-4V(ision) System Card,'' \emph{OpenAI Technical Report}, 2023. [Online]. Available: \url{https://cdn.openai.com/papers/GPTV_System_Card.pdf}

\bibitem{59}
W.~Dai \emph{et al.}, ``InstructBLIP: Towards general-purpose vision-language models with instruction tuning,'' \emph{arXiv preprint arXiv:2305.06500}, 2023.

\bibitem{60}
K.~Xuan \emph{et al.}, ``LEMMA: Towards LVLM-enhanced multimodal misinformation detection with external knowledge augmentation,'' \emph{arXiv preprint arXiv:2402.11943}, 2024.

\bibitem{61}
F.~Zeng \emph{et al.}, ``Multimodal misinformation detection by learning from synthetic data with multimodal LLMs,'' in \emph{Findings of ACL: EMNLP 2024}, pp.~10467--10484.

\bibitem{16}
F.~Monti \emph{et al.}, ``Fake news detection on social media using geometric deep learning,'' \emph{arXiv preprint arXiv:1902.06673}, 2019.

\bibitem{18}
Y.~Li \emph{et al.}, ``A survey on truth discovery,'' \emph{ACM SIGKDD Explorations}, vol.~17, no.~2, pp.~1--16, 2015.

\end{thebibliography}
\end{document}